\documentclass[a4paper,UKenglish,cleveref, autoref, thm-restate]{lipics-v2021}
\usepackage{pdflscape}
\usepackage{todonotes}
\usepackage{amsfonts} 
\usepackage{amsmath}
\usepackage{amssymb}
\usepackage{amsthm} 
\usepackage{mathtools} 
\usepackage{listings} 
\usepackage{subcaption}     
\usepackage{hyperref}
\usepackage{cleveref}
\crefformat{footnote}{#2\footnotemark[#1]#3}
\usepackage{xcolor}

\DeclareMathOperator*{\argmax}{arg\,max}

\hideLIPIcs  

\title{Improving deep learning precipitation nowcasting by using prior knowledge} 

\titlerunning{Improving deep learning precipitation nowcasting by using prior knowledge
} 

\author{Matej {Choma}}{Meteopress s.r.o, \and Faculty of Information Technology, Czech Technical University in Prague, Czech Republic }{chomamat@fit.cvut.cz	}{}{}

\author{Petr {Šimánek}}{Faculty of Information Technology, Czech Technical University in Prague, Czech Republic \and \url{https://fit.cvut.cz/en} 
}{petr.simanek@fit.cvut.cz}{[https://orcid.org/
0000-0001-5808-0865]}{}

\author{Jakub {Bartel}}{Meteopress s.r.o}{jakub.bartel@meteopress.cz}{}{}

\authorrunning{M. Choma and P. Šimánek and J. Bartel} 

\Copyright{Matej Choma and Petr Šimánek} 

\ccsdesc[500]{Computing methodologies~Machine learning algorithms}

\keywords{Nowcasting, spatio-temporal prediction, ConvLSTM, physics informed NN, PhyDNet} 

\category{} 

\relatedversion{} 

\nolinenumbers 

\EventEditors{John Q. Open and Joan R. Access}
\EventNoEds{2}
\EventLongTitle{29th International Symposium on Temporal Representation and Reasoning (TIME 2022)}
\EventShortTitle{TIME 2022}
\EventAcronym{TIME}
\EventYear{2022}
\EventDate{07--09 November, 2022}
\EventLocation{Online}
\EventLogo{}
\SeriesVolume{42}
\ArticleNo{23}

\begin{document}

\maketitle

\begin{abstract}
Deep learning methods dominate short-term high-resolution precipitation nowcasting in terms of prediction error. However, their operational usability is
limited by difficulties explaining dynamics behind the predictions, which are
smoothed out and missing the high-frequency features due to optimizing for
mean error loss functions. We experiment with hand-engineering of the advection-diffusion differential equation into a PhyCell to introduce more accurate physical prior to a PhyDNet model that disentangles physical and residual dynamics. Results indicate that while PhyCell can learn the intended dynamics, training of
PhyDNet remains driven by loss optimization, resulting in a model with the
same prediction capabilities.
\end{abstract}

\section{Introduction}
\label{sec:typesetting-summary}
 It is normal to adapt day-to-day activities with respect to temperature, wind, and precipitation outside. Homes are built as a shelter from the weather, and its effects on food production inspired cultures around the world. Thus, it is beneficial to know in advance what the weather may be like, and adjust according to it to increase comfort, safety, and profit. However, weather may sometimes be severe, changing in tens of minutes and destroying anything standing in its path. The tornado in Moravia, which happened on June 24, 2021, is a tragic example still in the living memory \cite{Pestova2021-he}. In these cases, weather prediction becomes a critical tool for protection.

Precipitation is not only dictating clothes, transport, or moisture for crops, but in our latitudes, it accompanies most of the short-term storm-based severe weather as well. Each time a dark cloud forms on the horizon, a question regarding its future development and severity arises. Luckily, precipitation may be monitored in real-time and in high resolution with weather radars. It may be argued that the observations are sufficient for taking individual protective measures. Nevertheless, humans have many activities when it is impossible to monitor their surroundings actively, and a localized short-term prediction may be game-changing.

We have been exploring the use of deep learning (DL) techniques for short-time high-resolution rainfall prediction in cooperation with the company Meteopress \cite{Choma2019-sl}. Building on the PhyDNet architecture disentangling physical from unknown dynamics \cite{Le_Guen2020-PhyDNet}, we have achieved unparalleled quantitative performance of an operational precipitation nowcasting system \cite{Choma2021-gr}. The difficulty of explaining dynamics learned by a DL model lowers the trustworthiness of the predictions in the eyes of meteorologists. The regression formulation of the learning problem, guided by mean error loss functions, results in the ignorance of hardly predictable high-frequency features, which are the most important ones during storm events. Last but not least, the performance decays quickly with prolonged forecast times.

PhyDNet is a neural network (NN) developed for a general video prediction, where the underlying dynamics governing the system are unknown. However, with the long history of weather forecasting \cite{Bauer2015-tm}, this is not the case for precipitation.
In this thesis, we aim to progress in addressing the issues mentioned above by exploiting the prior knowledge of precipitation physics. This work will explore how the human knowledge of the atmosphere may be used to enhance the physical part of the prediction in PhyDNet. Subsequently, models incorporating the proposed changes will be trained on a radar echo dataset and compared to a PhyDNet baseline. The results will be thoroughly analyzed and discussed.

\section{Related work}

Traditional multi-day weather forecasts are computed using numerical weather
prediction (NWP) models, which model physical atmospheric processes on a selected grid-scale as an initial value problem. Real-time high-resolution radar and satellite observations make accurate NWP initializations possible. However, the cost of data assimilation and limitations on the model resolution to maintain computability cause not an optimal use of this data for short-range $0-2$ h nowcasting. An accepted approach to this time range is to compute nowcasts as an extrapolation on a sequence of radar or satellite measurements. \cite{Prudden2020-qk}

In \textit{Lagrangian persistence} models, it is assumed that precipitation intensity does not change. An advection field (optical flow) is estimated from a sequence of past observations, and the future ones are predicted by advecting the present rainfall. An open-source library containing these models is \verb|rainymotion| \cite{Ayzel2019-zi}. There have been advances, building on the Lagrangian persistence, allowing probabilistic, more accurate nowcasts, such as models from the library \verb|pySTEPS| \cite{Pulkkinen2019-bt}. However, the nowcasting of convective initiation, development, and decay remains difficult. \cite{Prudden2020-qk}

``\textit{Machine learning provides an opportunity to capture complex non-linear spatio-temporal patterns and to combine heterogeneous data sources for use in prediction,}'' \cite{Prudden2020-qk}. The ConvLSTM architecture \cite{Shi2015-ConvLSTM} was initially designed for precipitation nowcasting, and improvements to spatio-temporal predictions were introduced in PredRNN \cite{Wang2021-iy}. A Deep Generative Model may be used to predict high-frequency features in the precipitation \cite{Ravuri2021-ce}.

\subsection{Physics and Deep Learning}
Enhancing DL models with a physics prior or a combination of physical modeling and DL can improve the ability of models to generalize to unseen samples, reduce the size of models or help training when not enough training data is available. A good overview of the topic may be found in \cite{Thuerey2021-wl}. The following work, alongside PhyDNet \cite{Le_Guen2020-PhyDNet}, influenced our research.
\begin{itemize}
	\item \textit{Physics-informed neural networks} \cite{Raissi2017-hu} are constrained by physical laws, expressed as general non-linear PDEs. These can learn solutions to supervised training problems data-efficiently while respecting the given laws.
	\item  In \cite{Raissi2018-pd} the authors present \textit{hidden fluid mechanics}, a DL framework for inference of hidden quantities, like fluid pressure and velocity, from spatio-temporal visualizations of a passive scalar. Passive scalar is transported by the fluid but has no dynamical effect on the fluid motion.
	\item APHYNITY \cite{Le_Guen2020-APHYNITY} is a framework for augmenting physical models with DL. The novel formulation of the learning problem allows the physical model to learn as much of the dynamics as possible.
\end{itemize}
\section{PhyDNet}\label{sec:3_disentangle}

PhyDNet \cite{Le_Guen2020-PhyDNet} is a recurrent NN (RNN) designed for a general prediction of future video frames that learns disentanglement between physical and unknown dynamics governing the system captured in the video. The approach proposed in \cite{Le_Guen2020-PhyDNet} builds on the idea of approximation of partial differential equations (PDEs) with convolutional filters and creates a way to include the equations in deep learning models. 



Given a frame of the video $\mathbf{u} ^{(t)} $ (for details about dimensions see Appendix \ref{appC2}), PhyDNet is trained to predict the following frame $\mathbf{u} ^{(t+\Delta)}$, under the assumption that the captured system can be at least partially described by some physical laws. The design of the architecture contains two branches. The first branch consists of PhyCell which models some differential operators and handles physical dynamics in the prediction. The second one is a deep ConvLSTM \cite{Shi2015-ConvLSTM} cell handling the residual dynamics. As the differential operators may not catch all the dynamics at the pixel level of the video, this disentanglement is preceded by an embedding to a latent space $\mathcal{H}$ that is learned end-to-end by deep convolutional encoder $E$ and decoder $D$. \cite{Le_Guen2020-PhyDNet}

PhyCell leverages physical prior to improve generalization and allows the model to learn some dynamics describable by PDE more effectively with less trainable parameters. ConvLSTM learns the complex unknown factors necessary for pixel-level prediction. \cite{Le_Guen2020-PhyDNet}

In the latent space $\mathcal H$, the memory of the PhyDNet cell stores learned embedding of a video up to a time $t$, in a domain with coordinates $\mathbf x=(x,y)$, represented as $\mathbf h(t, \mathbf x) = \mathbf h ^{(t)} \in \mathcal H$ and linearly disentangled into physical and residual components as $\mathbf h ^{(t)} = \mathbf h ^{(t)}_p + \mathbf h ^{(t)}_r$. Dynamics of the video are then governed by the following PDE:
\begin{equation}\label{eq:3_disentangle}
	\frac{\partial\mathbf h ^{(t)} }{\partial t}=
		\frac{\partial\mathbf h ^{(t)}_p }{\partial t}
		+ \frac{\partial\mathbf h ^{(t)}_r }{\partial t}
	\coloneqq \mathcal M_p(\mathbf h ^{(t)}_p, E(\mathbf u ^{(t)} ))
		+ \mathcal M_r(\mathbf h ^{(t)}_r, E(\mathbf u ^{(t)} )),
\end{equation}
where $\mathcal M_p$ is modeled by PhyCell and $\mathcal M_r$ by ConvLSTM.
Prediction of the next frame, discretized according to the forward Euler method, is computed as:
\begin{equation}\label{eq:3_disentangle_decoder}
	\widehat{\mathbf u} ^{(t+\Delta)} =D(\mathbf h ^{(t+\Delta)}_p + \mathbf h ^{(t+\Delta)}_r) = D(\mathbf h ^{(t)}_p + \mathcal M_p(\mathbf h ^{(t)}_p, E(\mathbf u ^{(t)} )) + \mathbf h ^{(t)}_r + \mathcal M_r(\mathbf h ^{(t)}_r, E(\mathbf u ^{(t)} ))),
\end{equation}
remembering the newly computed hidden states $\mathbf h ^{(t+\Delta)}_p $ and $ \mathbf h ^{(t+\Delta)}_r$. \cite{Le_Guen2020-PhyDNet}

\subsection{Physical Model -- PhyCell}

PhyCell is a novel ''physically constrained'' recurrent cell introduced in \cite{Le_Guen2020-PhyDNet} that models the dynamics in two steps:
\begin{equation}
	\mathcal M_p(\mathbf h _p, E(\mathbf u ))\coloneqq\Phi(\mathbf h_p)+C(\mathbf h_p, E(\mathbf u)).
\end{equation}
The first step is prediction in the latent space $\Phi(\mathbf h_p)$ (Equation \ref{eq:3_phi}) using a linear combination of spatial derivatives. Then, correction step $C(\mathbf h_p, E(\mathbf u))$ (Equation \ref{eq:3_correct}) handles the assimilation of input data into the latent representation similarly as in the Kalman filter \cite{Kalman1960-mn}.

\subsubsection{Prediction Step}\label{sec:3_phi}

The physical predictor $\Phi(\mathbf h_p)$ models a generic class of \textbf{linear} PDEs as
\begin{equation}\label{eq:3_phi}
	\Phi(\mathbf h_p ^{(t)} )\coloneqq
	\sum_{i,j<k}c_{i,j}\mathcal D_{i,j}(\mathbf h_p ^{(t)})
	=\sum_{i,j<k}c_{i,j}\frac{\partial^{i+j}\mathbf h_p}{\partial x^i\partial y^j}(t, \mathbf x) ,
\end{equation}
computing a linear combination of spatial derivatives using learned coefficients $c_{i,j}$. Following the forward Euler discretization, the latent prediction is computed as 
\begin{equation}
	\tilde{\mathbf h}_p ^{(t + \Delta)} = \mathbf h_p ^{(t)} + \Phi (\mathbf h_p ^{(t)}).
\end{equation}
As this step relies solely on the hidden representation $\mathbf h_p ^{(t)} $, prediction $\tilde{\mathbf h}_p ^{(t + \Delta)} $ can be computed  even if the input frame $\mathbf u ^{(t)} $ is not available. \cite{Le_Guen2020-PhyDNet}

The operation in Equation \ref{eq:3_phi} is implemented using two convolutional layers. The first one $\theta _1$ computes $k^2$ spatial derivatives as $\phi_1 = \theta_1 \circledast \mathbf h_p $, resulting in a tensor $\phi_1 \in \mathbb R^{k^2\times H_h \times W_h}$. This operation is described in detail in the Appendix \ref{appC1}. The second layer $\theta_2$ performs the linear combination as convolution $\phi_2=\theta_2 \circledast \phi_1$. The $C_h$ kernels of $\theta_2$ are sized $1\times 1$, assigning a scalar $c_{i,j}\in \mathbb R$ to each partial derivative, represented as a channel in $\phi_1$ and performing combination for each spatial position.


\subsubsection{Correction Step}

The correction step $C(\mathbf h_p, E(\mathbf u))$, guiding the assimilation of latent prediction and input data, is defined as
\begin{equation}\label{eq:3_correct}
	C(\mathbf h_p ^{(t)} , E(\mathbf u ^{(t)} )) \coloneqq \mathbf K ^{(t)}\odot 
	(E(\mathbf u ^{(t)} ) - (\mathbf h_p ^{(t)} + \Phi (\mathbf h_p ^{(t)}))) .
\end{equation}
Using this equation discretized according to forward Euler method, it is possible to derive the whole computation of the new hidden state of PhyCell as
\begin{equation}
\begin{split}
	\mathbf h_p ^{(t+\Delta)}&= \mathbf h ^{(t)} _p+\Phi(\mathbf h_p ^{(t)} )+C(\mathbf h_p ^{(t)} , E(\mathbf u ^{(t)} ))=\tilde{\mathbf h}_p ^{(t + \Delta)} + \mathbf K ^{(t)}\odot 	(E(\mathbf u ^{(t)}) - \tilde{\mathbf h}_p ^{(t + \Delta)})
	\\&= (1 - \mathbf K ^{(t)})\odot  \tilde{\mathbf h}_p ^{(t + \Delta)} + \mathbf K ^{(t)}\odot 	E(\mathbf u ^{(t)} ) .
\end{split}
\end{equation}
The gating factor $\mathbf K ^{(t)}\in [0,1]$ in these two equations, can be interpreted as a Kalman gain controlling the trade-off between the prediction and correction steps. When $\mathbf K ^{(t)}=0$, the input frame $\mathbf u ^{(t)} $ has no contribution to the computed hidden state $\mathbf h_p ^{(t+\Delta)}$. On the contrary, if $\mathbf K ^{(t)}=1$, the whole latent prediction $\tilde{\mathbf h}_p ^{(t + \Delta)}$ is discarded and the hidden state $\mathbf h_p ^{(t+\Delta)}$ is reseted according to the input. The Kalman gain $\mathbf K ^{(t)}$ is computed using two convolutional layers $\theta_3, \theta_4$ as $\mathbf K ^{(t)} = \theta_3\circledast\tilde{\mathbf h}_p ^{(t + \Delta)} + \theta_4\circledast E(\mathbf u ^{(t)} )$.

\subsection{Residual Model -- ConvLSTM} \label{sec:3_convlstm}
The unknown phenomena in the video dynamics that are not corresponding to prior models are learned entirely from the data as the residual dynamics $\mathcal M_r(\mathbf h _r, E(\mathbf u ))$ (Equation \ref{eq:3_disentangle}). This task is handled by a deep NN, such as ConvLSTM \cite{Shi2015-ConvLSTM}, which extends the idea of LSTM recurrent cell to spatio-temporal data and is used by the PhyDNet authors in \cite{Le_Guen2020-PhyDNet}. The residual model can also consider much more complex physical processes than the linear combination of partial differential equations. In particular importance for nowcasting modeling, we believe that ConvLSTM can learn to capture fat-tailed non-local (in time) effects that are common in storms and precipitations. These effects can be described by fractional calculus \cite{Jiang_2018}. The fractional time difference can not be readily computed by PhyCell and is hoped to be approximated by ConvLSTM. We will investigate this hypothesis in further work. Also, we will try to approximate fractional time derivatives directly in PhyCell with learned fractional order.   


\section{Problem Setup}\label{chapt:4}

Precipitation and storm nowcasting, formulated as a spatio-temporal prediction of atmospheric measurement sequences, poses an ideal problem for ML models, thanks to the large amounts of real-time, well-defined, and relatively clean data. This section formally defines the precipitation nowcasting problem explored and elaborates on the data and tools used.

\subsection{Nowcasting Problem Formulation}\label{sec:4_nowcasting}

We formulate the precipitation nowcasting task as a sequence to sequence prediction of tensors $\mathbf{\Psi}^{(T)}\in\mathbb{R}^{C\times H\times W}$, describing the state of the atmosphere at a given time $T$, with a constant time step $\Delta$. $\mathbf\Psi^{(T)} $ is a $3D$ tensor, where $H$ and $W$ are respectively the height and width of the prediction domain, and $C$ is the number of different data channels.

Given a sequence of past $\tau_I$ measurements $(\mathbf\Psi^{(T-(\tau_I - 1)\Delta)},\dots,\mathbf\Psi^{(T)})$ up to a time $T$, the task is to predict $\tau_O$ future ones as
\begin{gather}
\begin{split}
	&(\widehat{\mathbf\Psi}^{(T+\Delta)},\dots,\widehat{\mathbf\Psi}^{(T+\tau_O\Delta)})
	=\\
	&\argmax_{({\mathbf\Psi}^{(T+\Delta)},\dots,{\mathbf\Psi}^{(T+\tau_O\Delta)})}P(({\mathbf\Psi}^{(T+\Delta)},\dots,{\mathbf\Psi}^{(T+\tau_O\Delta)})|(\mathbf\Psi^{(T-(\tau_I - 1)\Delta)},\dots,\mathbf\Psi^{(T)})),
\end{split}\raisetag{2.7\baselineskip}
\end{gather}
where $\widehat{\mathbf\Psi}^{(t)}\in\mathbb{R}^{C_O\times H\times W}$ is a prediction of future precipitation fields for each timestamp $t \in (T+\Delta,\dots,T+\tau_O\Delta)$.  The number of input $C$ and output $C_O$ data channels may differ.

\subsection{Deep Learning Approach}\label{sec:4_dl_approach}

We approach this task using convolutional RNN $\mathcal F$ predicting the sequence of future states as a regression
\begin{equation}
	\mathcal F((\mathbf\Psi^{(T-(\tau_I - 1)\Delta)},\dots,\mathbf\Psi^{(T)}), \theta_F)=
	(\widehat{\mathbf\Psi}^{(T+\Delta)},\dots,\widehat{\mathbf\Psi}^{(T+\tau_O\Delta)}),
\end{equation}
where $\theta_F$ are trainable parameters of $\mathcal F$'s inner recurrent cell $F$. This cell is trained to predict the nearest future state as
\begin{equation}\label{eq:4_form_cell}
	F(\mathbf\Psi^{(T)}, \theta_F, \mathcal R_F)=\widehat{\mathbf\Psi}^{(T+\Delta)},
\end{equation}
where $\mathcal R_F$ is the cell's memory, which is updated after each use of the function. To predict $\tau_O$ future states the cell $F$ is used recurrently, processing the sequence chronologically. The prediction is discarded during the input states $(\mathbf\Psi^{(T-(\tau_I - 1)\Delta)},\dots,\mathbf\Psi^{(T-\Delta)})$, learning just the inner representation of the seen precipitation situation $\mathcal R_F$. The state $\widehat{\mathbf\Psi}^{(T+\Delta)}$ for the first lead time is predicted according to the Equation \ref{eq:4_form_cell}, and the following $\tau_O-1$ predictions are computed as
$	F(\widehat{\mathbf\Psi}^{(T+(i-1)\Delta)}, \theta_F, \mathcal R_F)=\widehat{\mathbf\Psi}^{(T+i\Delta)},$ for $i\in \{2,\dots,\tau_O\}$.

$\mathcal F$ is a supervised ML model, meaning that its parameters $\theta_F$ are learned on a dataset of $N$ training samples $\mathcal D_{Tr}=\{\mathbf X^{(i)}\}_{i=\{1:N\}}$. Each sample $$\mathbf X^{(i)}=(\mathbf\Psi^{(T-(\tau_I - 1)\Delta)},\dots,\mathbf\Psi^{(T)},\dots,{\mathbf\Psi}^{(T+\tau_O\Delta)})$$ is a sequence of $\tau_I+\tau_O$ atmospheric measurements, from which the first $\tau_I$ measurements $\mathbf X^{(i)}_I$ are used as an input to the model and the following $\tau_O$ as ground truth $\mathbf X^{(i)}_O$. Model parameters $\theta_F$ are learned through optimization of the loss function
\begin{equation}\label{eq:4_loss}
	\mathcal L(\theta_F, \mathcal D_{Tr})=\frac{1}{N}\sum_{i\in\{1:N\}}\left(\frac{1}{\tau_O}\sum _{j \in \{1:\tau_O\}} \mathcal L_{\text{step}}(\widehat{\mathbf\Psi}^{(T+j\Delta)}, \mathbf\Psi^{(T+j\Delta)})\right).
\end{equation}

\subsection{Radar Echo Dataset}
In this work, we consider a fixed dataset of radar echo image sequences. Thus, only $C=1$ input channel is used for each measurement $\mathbf\Psi^{(T)}$. The source radar echo data comes from the composite images created by and distributed through the OPERA program of EUMETNET \cite{opera_2016-xb}. 
At the time of the dataset creation, our archives contained 249176 images from the time window from 2015-10-23 19:30 UTC to 2020-07-21 23:50 UTC with a time step of 10 min. 
\begin{figure}
    \begin{center}
        \includegraphics[width=.5\textwidth]{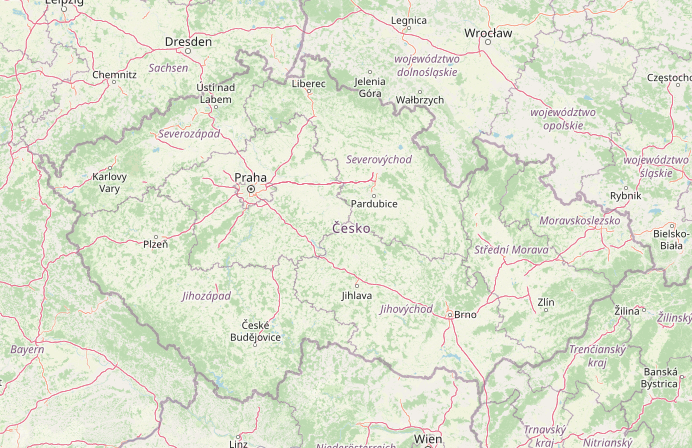}
    \end{center}
    \caption{The domain of the radar echo images in the dataset, visualized on OpenStreetMaps \cite{osm}.}
    \label{fig:dataset_crop}
\end{figure}

We decided to crop the prediction domain from the composite data to the area of the Czech Republic with small surroundings (Figure \ref{fig:dataset_crop}). Our reasoning behind this decision is to develop the models in an area with good radar coverage and quality measurements, keep the domain small enough for reasonable training times and memory requirements, and finally, a local preference. The WGS 84 coordinates\footnote{coordinates used in GPS, EPSG:4326 \cite{epsg}} of domain in degrees are
\begin{itemize}
	\item North-West corner -- 51.397001 N, 11.672296 E,
	\item South-East corner -- 48.223874 N, 19.274628 E.
\end{itemize}
The pictures show measurements above the Earth's surface in the Pseudo-Mercator geographic projection, known from the web mapping applications (projection code EPSG:3857 \cite{epsg}). We are aware of the potential pitfalls associated with Mercator projection that keeps local shapes but not distances. In the scope of this work, we do not identify the resolution difference as a problem. The intensity is preprocessed as described in Appendix \ref{appB}. The dataset is further split into training, validation, and testing sets as described in Appendix \ref{AppA}.


\section{PhyCell Adjustments for Precipitation Nowcasting}\label{chapt:5}

We propose changes to the PhyDNet's architecture, aiming at utilizing its strengths to the full potential in the context of nowcasting. PhyDNet is used as the recurrent cell $F$ in the Equation \ref{eq:4_form_cell}, where the cell's memory $\mathcal R_F=(\mathbf h_p, \mathbf h_r)$ contains hidden states of both physical and residual branches.

\subsection{Intensity Classification Loss}\label{sec:5_icloss}

One of the primary motivations for implementing precipitation nowcasting systems is the advantage of higher-resolution forecasts during storm events. Most short-time severe weather risks, such as hail, flash floods, strong winds, or lightning, are connected to convective systems and thus high-intensity precipitation. The ability to nowcast storms accurately, even if only dozens of minutes into the future, benefits both the general public and operational meteorologists monitoring these situations. The automatization of nowcasting brings unprecedented forecast localization to the end-users while providing another valuable information source for meteorologists issuing severe weather alerts.

However, when the training of a NN is formulated as a regression problem, the model may not be motivated to predict pixels with high intensities and other high-frequency features. Traditional regression loss functions, such as MSE, are penalizing an incorrect selection of the storm location twice -- once for predicting it at a wrong place and the second time for not predicting it at the correct one. To avoid these errors, regression-based models generally learn to express the uncertainty with smoothed-out predictions, omitting high-frequency features in the predictions.

While smoothed-out predictions achieve optimal prediction error, they do not provide enough information during storm events. To emphasize the prediction of high intensities, we propose to create a new output of the model, containing the prediction of ``probabilities'' of severe rainfall over 40 dBZ. According to Equation \ref{eq:4_form_cell}, output of one recurrent cell's prediction step is $\widehat{\mathbf\Psi}^{(T+\Delta)}$. In the case of the PhyDNet, it is the output of the deep convolutional decoder $D$ that is combining and processing predictions of the two branches (Equation \ref{eq:3_disentangle_decoder}). The new output is computed via a single convolutional layer $\theta_{\text{prob}}$ with a kernel size 3 as $
	\widehat{\mathbf\Psi}^{(T+\Delta)}_{\text{prob}} = \theta_{\text{prob}}\circledast \widehat{\mathbf\Psi}^{(T+\Delta)}.$ Considering these two outputs, the error of prediction on one training sample $\mathbf X ^{(i)}$ (Equation \ref{eq:4_loss}) decomposes to
\begin{equation}
	\mathcal L_{\text{sample}}(\mathcal F(\mathbf X^{(i)}_I, \theta_F), \mathbf X^{(i)}_O) = 
	\frac{1}{\tau_O}\sum _{j \in \{1:\tau_O\}} \mathcal L_{\text{img}}(\widehat{\mathbf\Psi}^{(T+j\Delta)}, \mathbf\Psi^{(T+j\Delta)})
	+ \mathcal L_{\text{icl}}(\widehat{\mathbf\Psi}^{(T+j\Delta)}_{\text{prob}}, \mathbf\Psi^{(T+j\Delta)}_{\text{prob}}).
\end{equation}
In this equation, the ground truth $\mathbf\Psi^{(T+j\Delta)}_{\text{prob}}$ is a binary image obtained via thresholding of the ground truth $\mathbf\Psi^{(T+j\Delta)}$ -- each pixel is assigned the value \textit{one} if its intensity is greater than 40 dBZ and \textit{zero} otherwise. The comparison of this binary ground truth and the predicted ``probabilities'' $\mathcal L_{\text{icl}}$ is called \textit{ICLoss} (Intensity Classification Loss) and performed via a cross-entropy loss as implemented in PyTorch\footnote{\url{https://pytorch.org/docs/stable/generated/torch.nn.CrossEntropyLoss.html}}. The loss is weighted towards the \textit{one} class with a scaling factor of 5 to reduce the imbalance of classes a bit.


\subsection{Non-linearity in the PhyCell}

A general aim of NNs is to be able to model a wide variety of functions on compact subsets of $\mathbb R^n$. As the perceptron function is linear, this universal function approximation trait is achieved by using a finite number of neurons in single or multiple layers with non-linear activation functions. \cite{Zhang2021-be}

Similar is true for CNNs such as the ConvLSTM (Section \ref{sec:3_convlstm}) used in the residual branch of the PhyDNet. In the default setting from \cite{Le_Guen2020-PhyDNet}, it is configured with three stacked cells, which are respectively operating on inputs with $(128, 128, 64)$ channels. While this setting theoretically gives it the ability to learn arbitrary functions, it is not designed to perform the point-wise multiplication of two images. PhyCell, with its linear prediction step, is not designed for multiplication as well (Equation \ref{eq:3_phi}). However, the multiplication of different physical quantities is a very common operation.

In Equation \ref{eq:3_phi}, scalars $c_{i,j}$ relevant to particular differential operators are learned during training and shared across all positions $\mathbf x$ of the domain. Considering the precipitation nowcasting task, the change of precipitation intensity at all times and all places of the domain would only be linearly dependent on the gradient of the hidden state. Referencing the non-linear advection term in Navier-Stokes equations for modeling fluids, we identify this linearity as a significant limitation for the correct precipitation prediction.

In the default PhyDNet, the parameter limiting the order of the partial derivatives computed is set as $k=7$. To the best of our knowledge, derivatives of this order are not used in NWP models \cite{Kalnay2003-ju}, usually, derivatives of only up to 2nd order are used. We believe that this setting reduces the potential of explainability of the predictions, which is a trait highly valued by meteorologists. Moreover, it creates a large space in the PhyCell for loss optimization, possibly reducing the robustness of PhyCell's predictions and interfering with the task of ConvLSTM. Thus, we use $k=3$ in our later experiments (limiting to second-order derivatives).

In the following subsections, we propose two different approaches to enable non-linearity in the physical prediction of PhyDNet. One approach is called Quadratic, which is not very successful and is described in Appendix \ref{AppQuad}.

\subsubsection{Advection-diffusion Equation}\label{sec:5_advdiff}
The other approach relies on hand-engineering of prior knowledge, using the \textit{advection-diffusion} PDE to model the precipitation in the PhyCell. The prediction step from the Equation \ref{eq:3_phi} theoretically changes to 
\begin{equation}\label{eq:5_advdiff}
	\Phi(\mathbf h_p ^{(t)} )= \underbrace{- c_0\frac{\partial\mathrm{u}_x\mathbf h_p}{\partial x}(t, \mathbf x)- c_1\frac{\partial\mathrm{u}_y\mathbf h_p}{\partial y}(t, \mathbf x)}_{\text{advection}}+\underbrace{c_{2}\frac{\partial^{2}\mathbf h_p}{\partial x^2}(t, \mathbf x)+c_3\frac{\partial^{2}\mathbf h_p}{\partial y^2}(t, \mathbf x)}_{\text{diffusion}},
\end{equation}
where $\mathbf u=(u_x, u_y)$ is a vector field by which the precipitation is advected. The original idea is inspired by the work on Hidden Fluid Mechanics by Raissi et al. \cite{Raissi2018-pd}. They assume that the flow of fluid is observed through a passive scalar that is moved by the flow described by Equation \ref{eq:5_advdiff} but not affecting it.

However, in the case of precipitation nowcasting, we do not aim to infer a global advection field, interpretable as wind, that would move precipitation as a passive scalar. We are rather interested in modeling precipitation local developments. Thus, based on the advection term of Navier-Stokes equations (as used in \cite{Thuerey2021-wl}), $\mathbf u^{(t)} $ is inferred  from the system state $\mathbf h_p ^{(t)} $, guided just by the use of $\mathbf u^{(t)} $ in Equation \ref{eq:5_advdiff}. This approach introduces non-linearity to the PhyCell as $\mathbf u^{(t)} $ is a function of $\mathbf h_p ^{(t)} $.

The advection vectors $\mathbf u^{(t)} $ at time $t$ are computed with a single convolutional layer $\theta_U$ with kernel size $5$ as
\begin{equation}\label{eq:5_u}
	\mathbf u ^{(t)}  = U(\mathbf h_p ^{(t)} ) = \theta_U\circledast \mathbf h_p ^{(t)}.
\end{equation}
Following the original implementation of PhyCell, partial derivatives are computed with learned differential operators $\mathcal D_{i,j}$. Thus, four terms of the implemented PDE are\footnote{Omitting the time index $^{(t)}$ for clarity.}
\begin{equation}
\begin{split}
	\mathbf d (\mathbf h_p ) = \left(\mathcal D_{1,0}\left(U(\mathbf h_p )_x\mathbf h_p \right), \mathcal D_{0,1}\left(U(\mathbf h_p )_y\mathbf h_p\right), 
	\mathcal D_{2,0}\left(\mathbf h_p\right), \mathcal D_{0,2}\left(\mathbf h_p\right)
	\right),
\end{split}
\end{equation}
which are linearly combined using coefficients $\mathbf c = (c_0, \dots, c_3)$, learned through $1\times 1$ convolution.

Following problems with convergence during training, we have added Group Normalization\footnote{\url{https://pytorch.org/docs/stable/generated/torch.nn.GroupNorm.html}} ($GN$) to standardize the equation terms as one group. While we are struggling with the interpretation of $GN$ in terms of physical simulation, it should be noted that the original implementation of PhyCell\footnote{\label{foot:pdn}\url{https://github.com/vincent-leguen/PhyDNet}} uses $GN$ on the partial derivatives as well, splitting the 49 terms into 7 groups. Finally, the prediction step is implemented as
\begin{equation}
	\Phi(\mathbf h_p ^{(t)} ) = \mathbf c \cdot GN(\mathbf d(\mathbf h_p ^{(t)} )).
\end{equation}

\section{Experiments}\label{chapt:6}

This section summarizes concluded experiments, results, and findings \mbox{acquired} during PhyDNet adjusting for precipitation nowcasting. 
\subsection{Intensity Classification Loss}

\begin{figure}
     \centering
     \begin{subfigure}[b]{0.45\textwidth}
         \centering
         \includegraphics[width=1.0\textwidth]{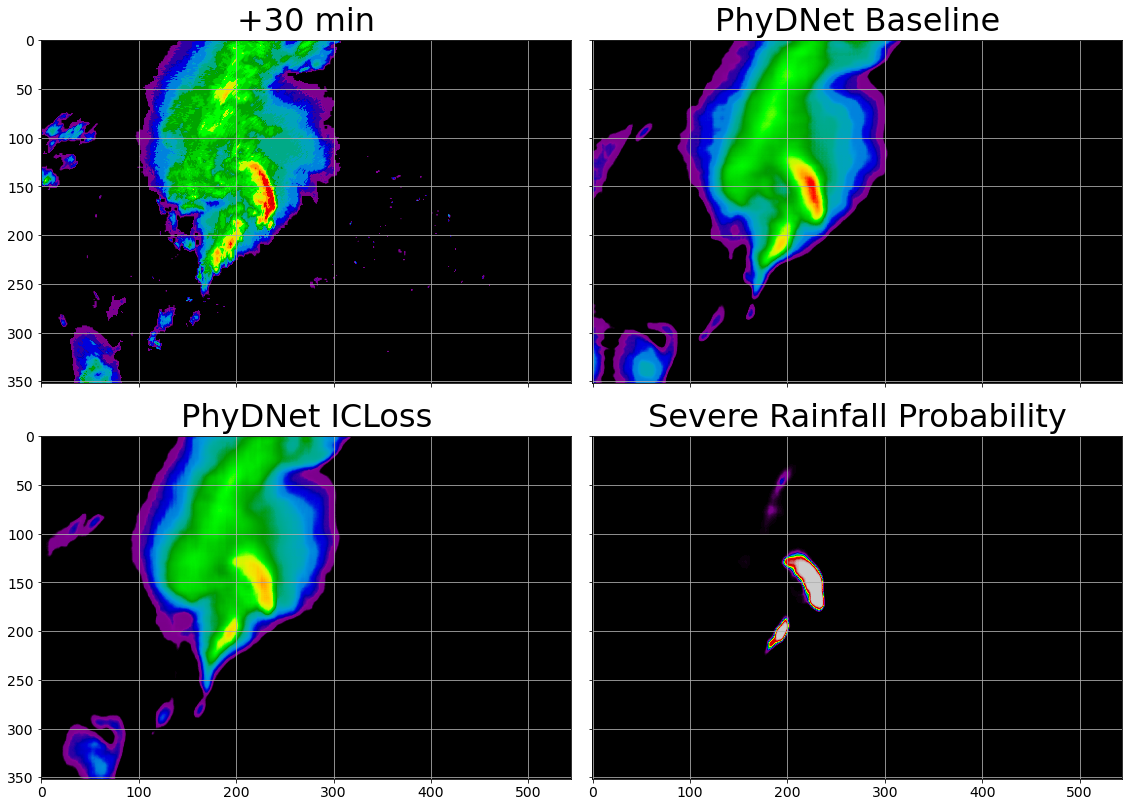}
         \caption{Effect of training with \textit{ICLoss} on prediction for 30 min (test set). The top left image is ground truth.}
         \label{fig:t11954_icloss_baseline}
     \end{subfigure}
     \hfill
     \begin{subfigure}[b]{0.45\textwidth}
         \centering
         \includegraphics[width=1.0\textwidth]{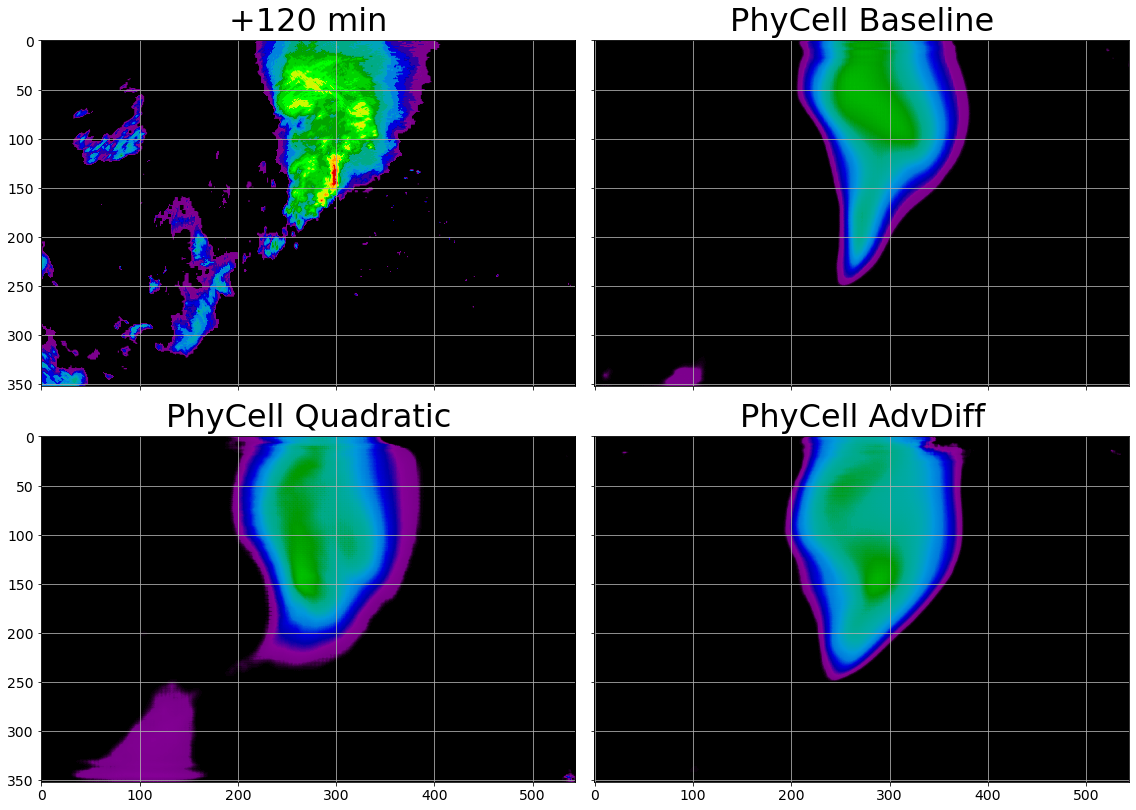}
         \caption{Sample prediction by pure PhyCell with different designs of $\Phi$ for 120 min (test set). The top left image is ground truth.}
         \label{fig:t11954_phycells_120}
     \end{subfigure}
     \hfill
     \caption{Sample effects of changes in PhyDNet.}
        \label{fig:three graphs}
\end{figure}

The effects of \textit{ICLoss} (Section \ref{sec:5_icloss}) were studied during the development phase on the validation set and \verb|PhyDNet Baseline| model. Its effects are well seen in Figure \ref{fig:t11954_icloss_baseline}, where the baseline model concentrates on precipitation with a larger area but smaller intensity, not predicting the small-area storms at all. On the other hand, \verb|PhyDNet ICLoss| does a better job in the identification of the storms and does not smooth them out, while the ``probability'' output correctly marks at least the lower part of the storms.

Results summarized in Table \ref{tab:exp_m_icloss} in terms of relative changes to performance, there is almost no difference in achieved MAE, MSE, or SSIM. However, the trade-offs of training with \textit{ICLoss} can be seen in other metrics. The decrease in the low-threshold CSI alongside better high-threshold CSI and the slight performance improvement over time, which may be seen in the plots, support the impression that predictions are less smoothed out if \textit{ICLoss} is used. The improvement in Kolmogorov-Smirnov distance suggests that focus on intensities over 40 dBZ makes empirical CDFs of the predictions more similar to the ground truth.

\begin{table}[h]
    \caption{Relative change in the metrics of PhyDNet ICLoss and PhyDNet AdvectionDiffusion compared to PhyDNet Baseline, PhyCell Quadratic, PhyCell AdvectionDiffusion compared to PhyCell Baseline (red denotes performance loss).}
    \label{tab:exp_m_icloss}
    \centering
    \begin{tabular}{|p{0.15\linewidth}||>{\raggedright\arraybackslash}p{0.15\linewidth}|>{\raggedright\arraybackslash}p{0.15\linewidth}|>{\raggedright\arraybackslash}p{0.15\linewidth}|>{\raggedright\arraybackslash}p{0.15\linewidth}|}
        \hline
        \textbf{Metrics} & \textbf{ICLoss} & \textbf{PhyCell Quad} & \textbf{PhyCell AD } & \textbf{PhyDNet AD}  \\
        \hline
        \hline

        \textbf{CSI 8 dBZ}          & $\color{red}-2.30\ \%$ & $\color{red}-3.72\ \%$    & $\color{red}-1.81\ \%$ & $\color{red}-3.41\ \%$ \\\hline
        \textbf{CSI 40 dBZ}         & $+6.46\ \%$ & $\color{red}-9.41\ \%$    & $+2.70\ \%$ & $\color{red}-6.14\ \%$  \\\hline
        \textbf{MAE}                & $\color{red}+0.25\ \%$     & $\color{red}+2.34\ \%$    & $\color{red}+1.00\ \%$ & $\color{red}+0.14\ \%$ \\\hline
        \textbf{MSE}                & $-0.38\ \%$   & $\color{red}+4.29\ \%$    & $\color{red}+2.46\ \%$  & $-0.40\ \%$   \\\hline
        \textbf{Kol-Smir} & $-2.13\ \%$  & $\color{red}+16.26\ \%$    & $\color{red}+9.73\ \%$ & $\color{red}+9.82\ \%$   \\\hline
        \textbf{SSIM}               & $\color{red}-0.22\ \%$   & $\color{red}-0.70\ \%$    & $+0.12\ \%$   & $\color{red}-0.11\ \%$  \\\hline
    \end{tabular}
    \end{table}

    
Following these observations, we have decided to use \textit{ICLoss} in our later experiments. However, closer inspection of its effects on the sample predictions from the test set shows limitations of the proposed \textit{ICLoss} implementation. Due to the setting of the threshold to 40 dBZ, the model tends to quickly lower the predicted intensities to this value. This may be clearly seen in the Figure \ref{fig:t11954_icloss_baseline}, but it happens in the first example as well. Moreover, due to the small capacity of the convolutional module producing the ``probabilities'' (Section \ref{sec:5_icloss}), these outputs lack gradient and very closely resemble the predicted intensities to be truly interpreted as probabilities of severe rainfall.
\subsection{Non-linearity in the PhyCell}\label{sec:6_phycell}

The predictions possibly learned by PhyCell with various designs of the prediction step $\Phi$ (Equation \ref{eq:3_phi}), are studied using PhyDNet models without the deep ConvLSTM residual branch. Firstly, the designs differ by the number of terms in $\Phi$.
\begin{itemize}

	\item \verb|PhyCell Baseline| computes linear combination of 49 spatial partial derivatives $\mathcal D_{i,j}(\mathbf h_p ^{(t)})$ for $i,j<7$.
	\item \verb|PhyCell Quad| combines 9 of the first-degree terms $\mathcal D_{i,j}(\mathbf h_p ^{(t)})$ for $i,j<3$, with all 45 of their possible second-degree combinations for a total of 54 terms.
	\item  \verb|PhyCell AdvDiff| utilizes only differential operators $(\mathcal D_{0,1}, \mathcal D_{1,0}, \mathcal D_{0,2}, \mathcal D_{2,0})$, having 4 terms in $\Phi$, out of which two are non-linear.
\end{itemize}

\begin{figure}
     \centering
     \begin{subfigure}[b]{0.45\textwidth}
         \centering
         \includegraphics[width=1.0\textwidth]{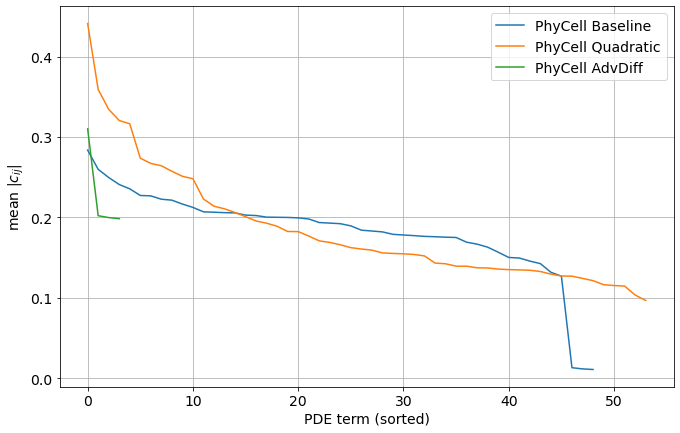}
         \caption{Mean absolute value of $c_{i,j}$ linear combination coefficients of PhyCell predictions step.}
         \label{fig:exp_amp_cij}
     \end{subfigure}
     \hfill
     \begin{subfigure}[b]{0.45\textwidth}
         \centering
         \includegraphics[width=1.0\textwidth]{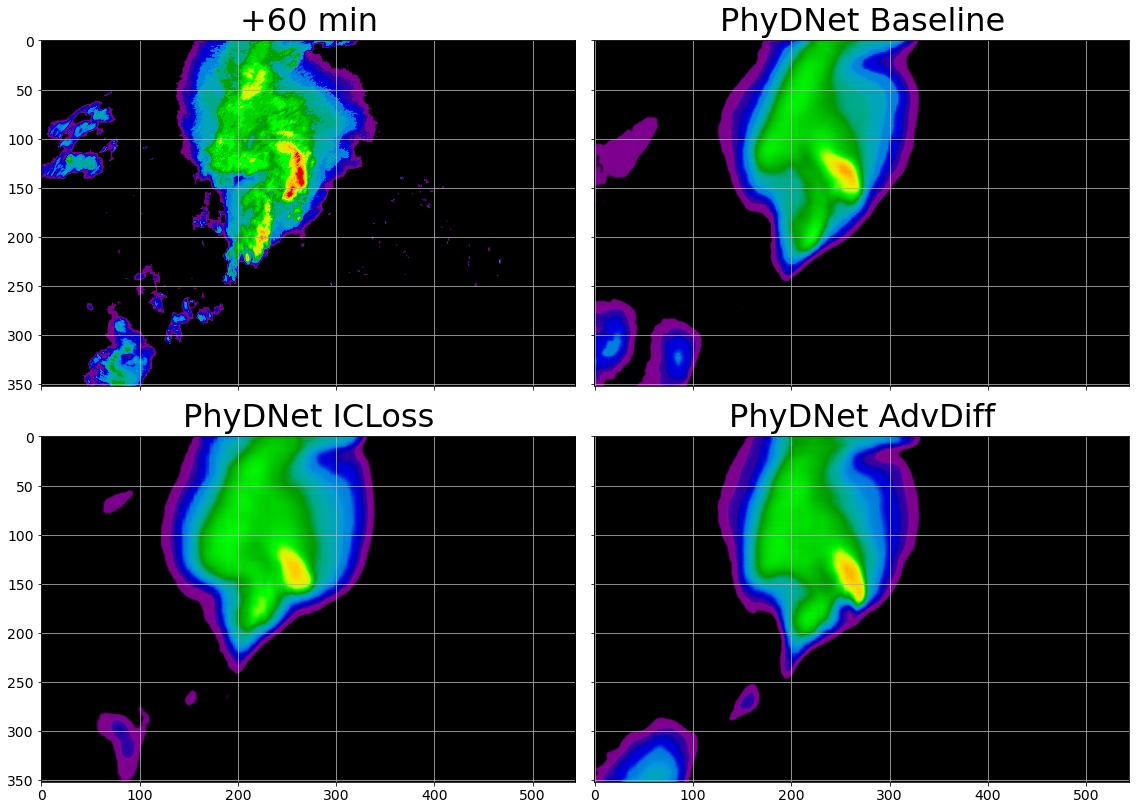}
         \caption{Sample prediction of convective precipitation by PhyDNet versions for 60 min (test set). The top left image is ground truth.}
         \label{fig:t11954_phydnets}
     \end{subfigure}
     \hfill
     \caption{}
\end{figure}

Consequently, \verb|PhyCell AdvDiff| has significantly less capacity to encode precipitation dynamics than the other two. The utilization of these terms visualized through $c$ coefficients of $\Phi$ in Figure \ref{fig:exp_amp_cij} shows that \verb|PhyCell Quad| prioritizes some terms more than others when compared to \verb|PhyCell Baseline|. The Top 10 utilized terms by \verb|PhyCell Quad| expressed in terms of used differential operators are
\begin{equation*}
	\left(\mathcal D_{0, 0}, \mathcal D_{1, 0}, \mathcal D_{0, 1}, \mathcal D_{1, 2}, \mathcal D_{2, 0}, \mathcal D_{0, 2}, \mathcal D_{1, 1},
	 (\mathcal D_{0, 0}\cdot\mathcal D_{1,0}), (\mathcal D_{0, 0}\cdot\mathcal D_{1,1}), (\mathcal D_{0, 0}\cdot\mathcal D_{1,2})\right).
\end{equation*}
The fact that these are either of first degree or multiplied with an undifferentiated hidden state ($\mathcal D_{0,0}$) hints that this implementation of non-linearity in $\Phi$ is not effective. Moreover, both sample predictions (Figure \ref{fig:t11954_phycells_120}) and quantitative evaluations (Table  \ref{tab:exp_m_icloss}) do not show any interesting results.

As summarized in Table \ref{tab:exp_m_icloss}, the proposed designs of PhyCell have worse quantitative performance when compared to the baseline. However, the aim of PhyCell is to give physically sound predictions on which the residual ConvLSTM module can build (Section \ref{sec:3_disentangle}) rather than achieve the best possible quantitative performance alone. Thus, the results of \verb|PhyCell AdvDiff|, given its much smaller capacity of $\Phi$, indicate that it learns precipitation dynamics much more effectively in terms of model size. It may be seen from the sample predictions that \verb|PhyCell Baseline| tends to smooth out the outputs to optimize for the loss, which is not the case for \verb|PhyCell AdvDiff|. The sample in Figure \ref{fig:exp_amp_cij}, containing predictions for twice the length of the training horizon, shows that \verb|PhyCell AdvDiff| is the only model to correctly predict the position of the high-intensity precipitation this far into the future, even though without the correct intensities. This hypothesis of less smoothed predictions that are better at predicting the location of the phenomena is supported by the gain in the high-threshold CSI alongside decay in the low-threshold one.
\subsection{Evaluation of PhyDNet AdvDiff}

In this section, \verb|PhyDNet AdvDiff| is compared to the \verb|PhyDNet Baseline| to evaluate the overall effect of the proposed changes on the prediction performance. \verb|PhyDNet ICLoss| model is included, to distinguish between the changes introduced by \textit{ICLoss} and the \textit{advection-diffusion} equation in the PhyCell. As summarized in Table \ref{tab:exp_m_icloss}, all three models achieve very similar mean errors, and their specifics are projected into trade-offs in other metrics. However, unlike in the previous section, sample predictions on the test set subjectively do not show any features that would clearly differentiate them (an example prediction of convective precipitation in Figure \ref{fig:t11954_phydnets} and of stratiform precipitation in Figure \ref{fig:t5020_phydnets}).

\begin{figure}
    \begin{center}
        \makebox[\textwidth][c]{\includegraphics[width=0.6\textwidth]{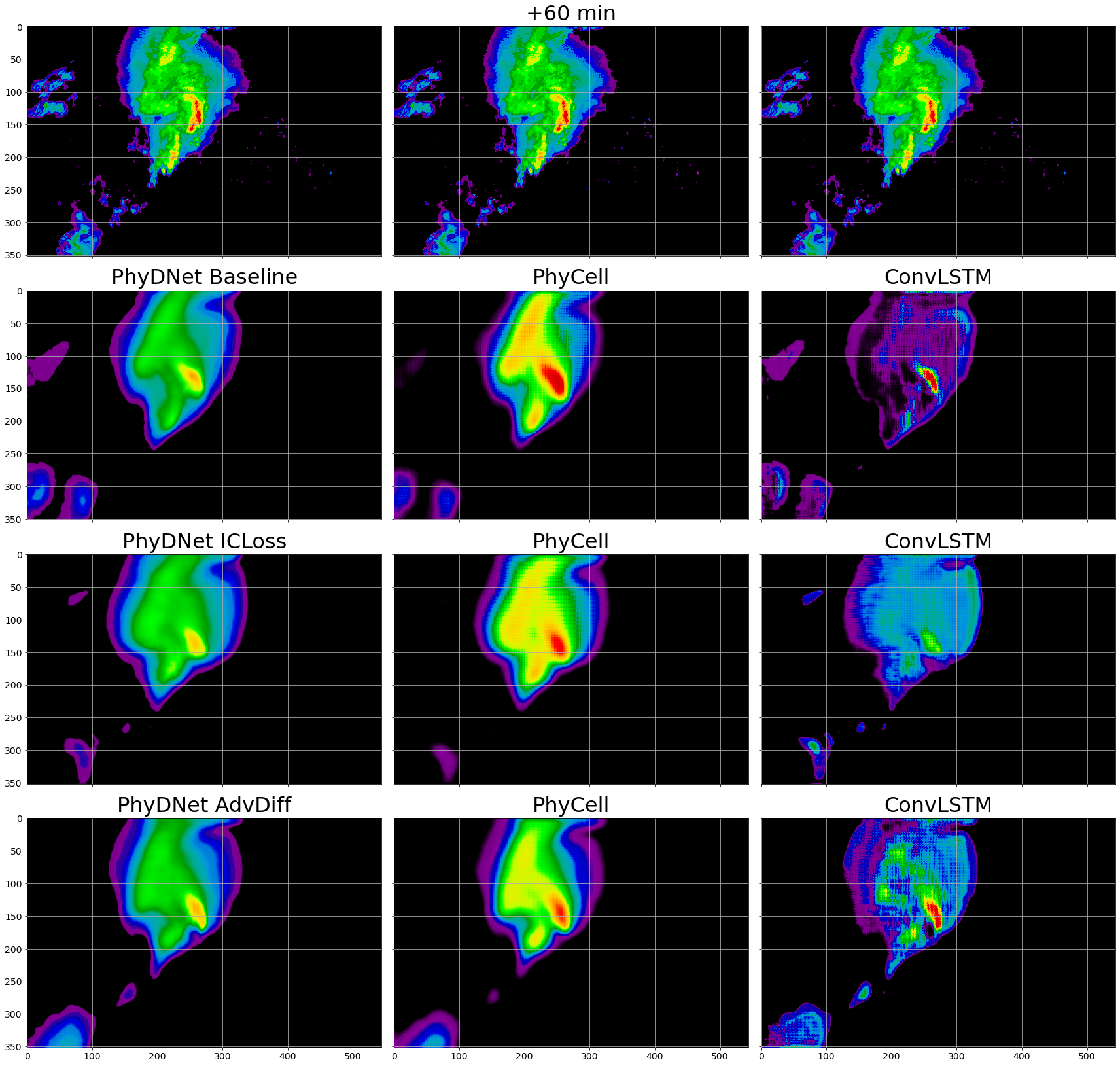}}
    \end{center}
    \caption{Decomposition of PhyDNet branches on a prediction for 60 min (test set). The top row displays the ground truth three times.}
    \label{fig:t11954_phydnets_dec}
\end{figure}

A difference among the predictions may be observed if predictions are decomposed into physical and residual branches, which are separately reconstructed through decoder $D$ and visualized (Figure \ref{fig:t11954_phydnets_dec}).
ConvLSTM of the \verb|PhyDNet AdvDiff| learns predictions containing a variety of objects and intensities. In contrast, the ConvLSTM of \verb|PhyDNet Baseline| predicts only objects with high intensities, and the predictions of \verb|PhyDNet ICLoss| ConvLSTM lack structure altogether. Thus subjectively, \verb|PhyDNet AdvDiff| utilizes the residual part the most. To partially quantify this hypothesis, Figure \ref{fig:test_mae} presents values of MAE for PhyCell and PhyDNet in one plot. While there is a difference in PhyCell errors, there is almost none in the case of PhyDNet -- in different models, ConvLSTM contributed different amounts to the overall performance. 

The local advection field is discussed in Appendix \ref{appAdvection}.


\section{Conclusion}

There is a difference in the learned dynamics of \textit{advection-diffusion} PhyCell when trained alone and as a part of PhyDNet, alongside the different amount of contribution to the prediction by ConvLSTM (Figure \ref{fig:test_mae}). These results lead us to speculate that only a limited amount of dynamics governing the precipitation are learnable under the current problem set. Results in Section \ref{sec:6_phycell} indicate that PhyCell can utilize the provided physics prior to predicting precipitation. However, combining PhyCell with a high-capacity ConvLSTM and training it for regression using a mean error loss function neglects the physics learned by the regularized PhyCell, in favor of smoothed predictions with unchanged performance. Moreover, the varying contribution of ConvLSTM hints that it can possibly learn more complex dynamics, given a change in the training formulation. This observation points to the loss function selection as the main limitation of the current approach.

However, there are caveats to this theory that need to be addressed. Firstly, trying to infer a global advection field as well, we experimented with $\mathbf u ^{(t)} $ computation from multiple previous states and with enforcing more physics through a continuity equation both resulting in divergence during the training. Two possible reasons are that the convolutional module $U$ (Equation \ref{eq:5_u}) does not have enough capacity to capture this complicated vector field or that there is not enough input data for its inference (e.g., it rains only in some parts of the domain). Secondly, the training of disentanglement in PhyDNet was not studied sufficiently. It may be possible that pre-training of some modules, introducing non-linear disentanglement, or emphasizing predictions of the PhyCell over the ones of ConvLSTM could lead to different results.
\clearpage

%
%
%
%
%
%



\bibliography{text_mybibliographyfile.bib}

\appendix
\section{Dataset Splitting}\label{AppA}
As a general practice, we split the dataset into \textit{train, validation}, and \textit{test}, using them isolated in various steps of the model development.

With time-series data, such as radar echo sequences, two consecutive samples capture the same atmospheric situation, just slightly shifted. Thus, the samples cannot be split randomly, as memorizing one sample will help predict the following one. The inclusion of these in different datasets would create a false performance. We decided to identify continuous precipitation situations, independent among themselves, and split them randomly into sets.

Two consecutive precipitation situations are considered independent if they are separated by at least 24 hours without any rainy radar echo measurements. The time delta of 24 hours was chosen empirically to ensure that memorizing a sample from one situation will not help with the prediction from a different one.

We have identified 275 independent precipitation situations in the considered time range with a median length of 78 hours and a mean length of 112 hours. These situations were split randomly into the train, validation, and test set with ratios and counts summarized in Table \ref{tab:dataset_split}.

\begin{table}[h]
\caption{Dataset statistics.}
\label{tab:dataset_split}
\centering
\begin{tabular}{|p{0.13\linewidth}||>{\raggedright}p{0.23\linewidth}|>{\raggedright}p{0.23\linewidth}|>{\raggedright\arraybackslash}p{0.23\linewidth}|}
\hline
\textbf{Dataset} & \textbf{Situation split percentage} & \textbf{\# independent situations} & \textbf{Hours of precipitation}          \\
\hline\hline
Train       & $72\%$       & 198  & 22724   \\\hline
Validation  & $\sim12.7\%$ & 35   & 3678    \\\hline
Test        & $\sim15.3\%$ & 42   & 4570    \\\hline
\end{tabular}
\end{table}

Each independent precipitation situation $\mathcal S_{T_a}^{T_b}$ consists of a set of training sequences $\{\mathbf X(t) \}_{t\in [T_a, T_b]}$, which are added to the corresponding dataset $\mathcal D$. The notation $\mathbf X(t) $ stands here for a sequence centered around the measurement $\mathbf\Psi^{(t)}$ as $$\mathbf X(t)=(\mathbf\Psi^{(t-(\tau_I - 1)\Delta)},\dots,\mathbf\Psi^{(t)},\dots,{\mathbf\Psi}^{(t+\tau_O\Delta)}).$$

\section{Precipitation Intensity}\label{appB}

The precipitation intensity is represented in the source data with 8-bit values using the \textit{dBZ} units. The scale is linearly mapping values $[0, 60]\ \mathrm{dBZ}$ to integers in $[0, 255]$, except the $0\ \mathrm{dBZ}$ measurement rendered as \textit{no precipitation}. Any measurement from range $(-\infty, 0]\ \mathrm{dBZ}$ is represented as the 0 value. We have chosen to linearly scale the 8-bit integers to float numbers in the range $[0,1]$, internally called \textit{MLdBZ}.

For a domain of the size of the Czech Republic, it is not raining every day. Thus, not every radar echo image contains information valuable for the ML model training, and \textit{rainy} images need to be selected.
\clearpage
\begin{definition}
	A precipitation field $\mathbf\Psi^{(T)}$ is flagged as \textbf{rainy} if
\begin{itemize}
	\item $>7\%$ of its area contains non-zero values,
	\item or $>1\%$ of its area has values $>24\ \mathrm{dBZ}$.
\end{itemize}
\end{definition}

After removing clearly noisy samples, we have identified 102873 rainy radar echo measurements in the studied time range.

\section{Details of PhyDNet Architecture}\label{appC}

\subsection{Approximation of Partial Derivatives}\label{appC1}

The partial differential operators $\mathcal D_{i,j}(\mathbf h_p)=\frac{\partial^{i+j}\mathbf h_p}{\partial x^i\partial y^j} = q_{i,j}\circledast \mathbf h_p$ are learned through constrained convolutional kernels $q_{i,j}$. The $k\times k$ moment matrix $\mathbf M(q_{i,j}) = (m_{a,b})_{k\times k}$ of a $k\times k$ convolutional kernel $q_{i,j}$ is defined as
\begin{equation}
	m_{a,b} \coloneqq \frac{1}{a!b!}\sum_{u,v=-\frac{k-1}{2}}^{\frac{k-1}{2}}u^a v^b q_{i,j}[u, v].
\end{equation}
It is shown in \cite{Le_Guen2020-PhyDNet} that if $m_{a,b}= 1$ for $a = i, b=j$ and $m_{a,b}=0$ otherwise, the convolutional kernel $q_{i,j}$ approximates differential operator $\mathcal D_{i,j}$ through finite difference coefficients.

Thus, the correct kernels are learned through $\mathcal L_m$ moment loss regularization. Defining a $k\times k$ matrix $\mathbf \Delta^k_{i,j}$, which equals 1 at position $(i,j)$ and 0 elsewhere, the regularization term is computed as
\begin{equation}
	\mathcal L_m = \sum_{i,j\leq k}||\mathbf M(q_{i,j}) - \mathbf \Delta^k_{i,j}||_F,
\end{equation}
where $||\cdot||_F$ is Frobenius norm.
\cite{Le_Guen2020-PhyDNet}

\subsection{Data Dimensions}\label{appC2}

The input video frames are $\mathbf u ^{(t)} \in \mathbb R^{C_u\times H \times W}$, where $H, W$ describe size of the frame and $C_u$ is number of input channels (typically $C_u \in \{1, 3, 4\}$). The dimensions change after the learned embedding into the latent space to $E(\mathbf u ^{(t)} ) \in \mathcal H = \mathbb R^{C_h\times H_h\times W_h}$. In the default settings $C_h = 64$, height of the transformed domain is $H_h=H / 4$ and width $W_h = W/ 4$. PhyDNet design is fully convolutional; thus, it can handle input images with arbitrary $H$ and $W$.

\begin{figure}
    \begin{center}
        \makebox[\textwidth][c]{\includegraphics[width=0.6\textwidth]{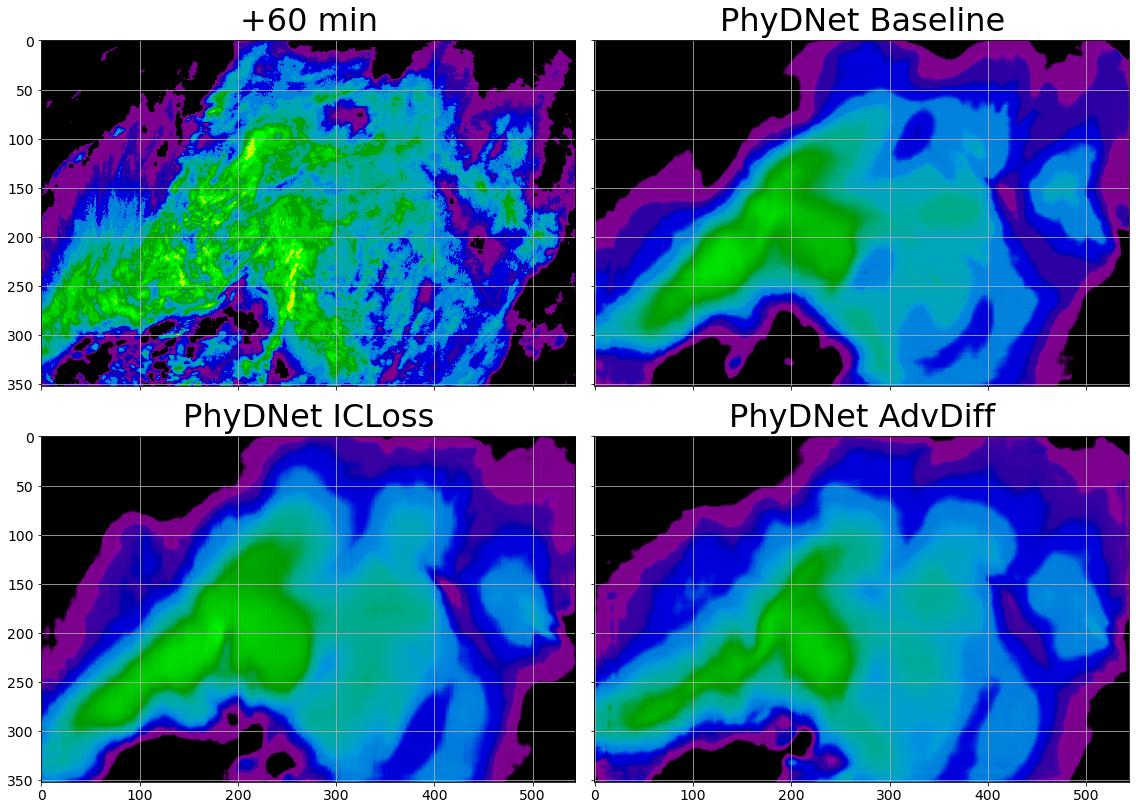}}
    \end{center}
    \caption{Sample prediction of stratiform precipitation by PhyDNet versions for 60 min (test set). The top left image is ground truth.}
    \label{fig:t5020_phydnets}
\end{figure}

\begin{figure}
	\vspace{-15pt}
    \begin{center}
        \makebox[\textwidth][c]{\includegraphics[width=0.8\textwidth]{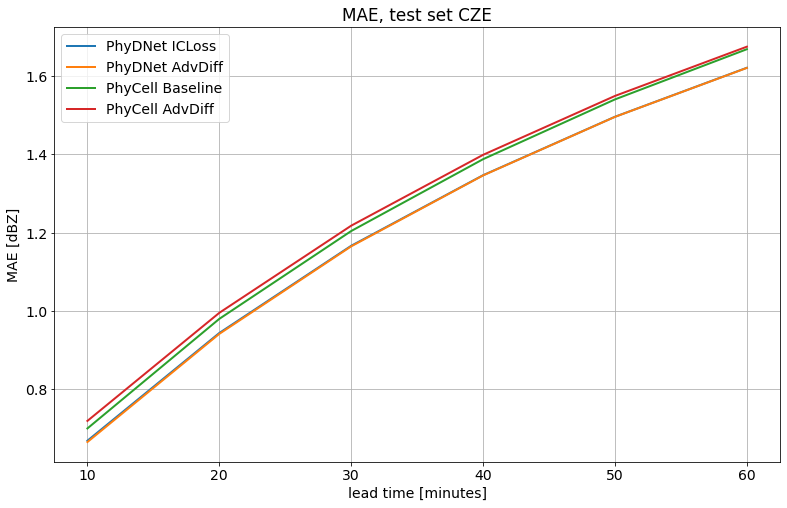}}
    \end{center}
    \caption{MAE on the test set.}
    \label{fig:test_mae}
    \vspace{-15pt}
\end{figure}

\section{Advection Field Inferred by PhyCell AdvDiff}\label{appAdvection}

\begin{figure}
    \begin{subfigure}{\textwidth}
        \centering
        \makebox[\textwidth][c]{\includegraphics[width=0.8\linewidth]{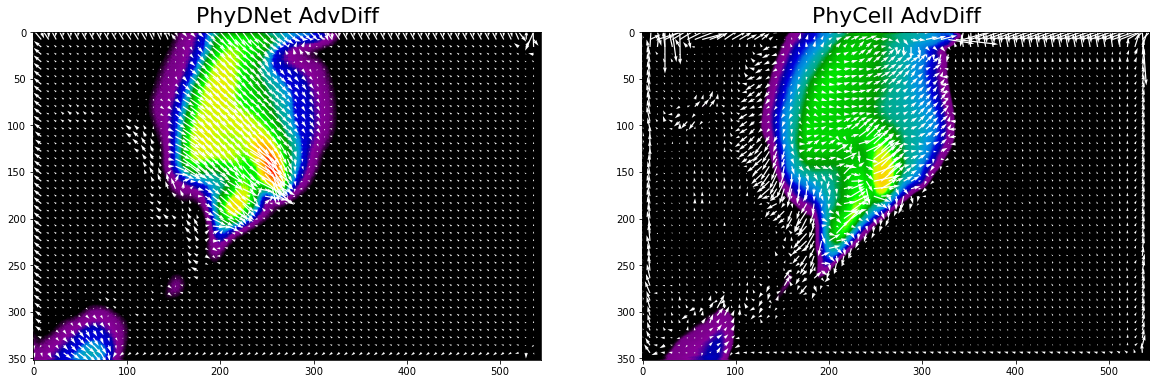}}
        \caption{convective precipitation moving East-North-East}
        \label{fig:t11954_velocity}
    \end{subfigure}
    \begin{subfigure}{\textwidth}
        \centering
        \makebox[\textwidth][c]{\includegraphics[width=0.8\linewidth]{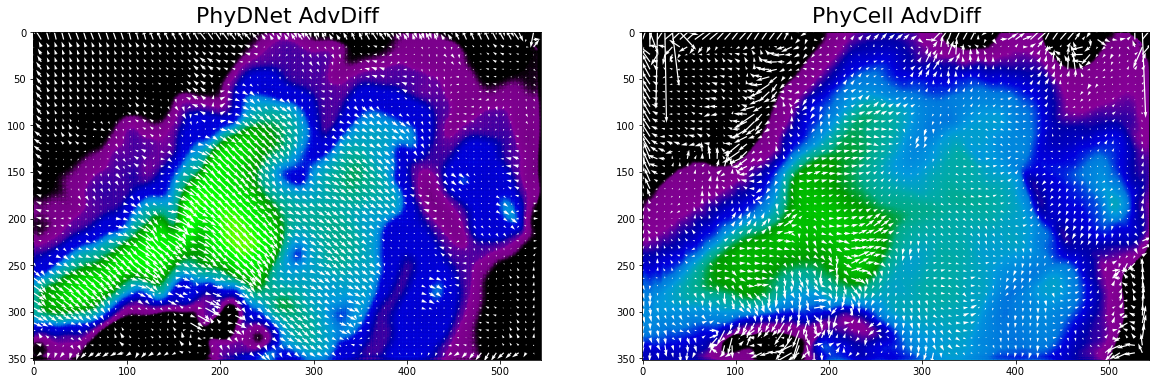}}
        \caption{stratiform precipitation moving East}
        \label{fig:t5020_velocity}
    \end{subfigure}
    \begin{subfigure}{\textwidth}
        \centering
        \makebox[\textwidth][c]{\includegraphics[width=0.8\linewidth]{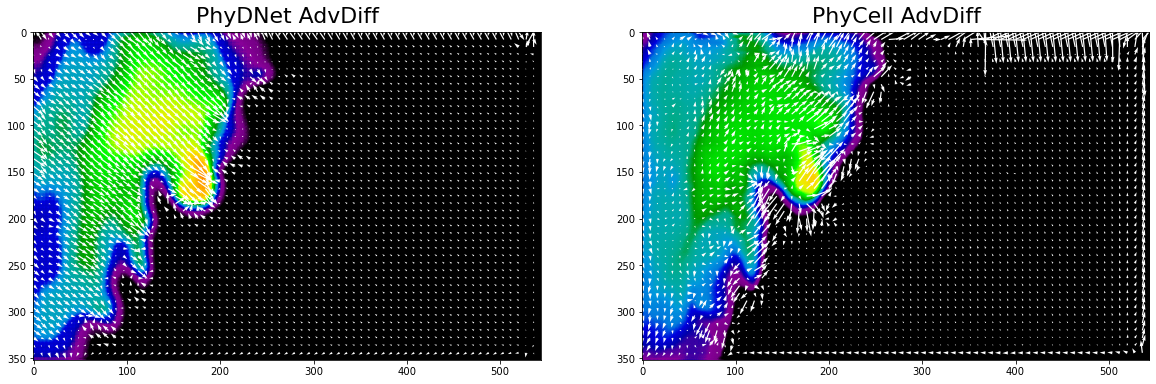}}
        \caption{convective precipitation moving North-North-East}
        \label{fig:t2364_velocity}
    \end{subfigure}
    \caption{Advection field $\mathbf u$ plotted over a PhyCell partial prediction for 60 min (test set).}
    \label{fig:6_velocity}
\end{figure}

%
This section contains visualizations (Figure \ref{fig:6_velocity}) of the advection field $\mathbf u ^{(t)} $, inferred from the observed data through their use in the \textit{advection-diffusion} equation in PhyCell (Section \ref{sec:5_advdiff}). $\mathbf u ^{(t)} $ is computed from a single hidden state $\mathbf h_p ^{(t)}$, which should theoretically contain complete information about the current precipitation situation. In the case of \verb|PhyCell AdvDiff|, it may be observed that direction of the vectors matters. However, instead of the general motion vectors (interpretable as wind), these rather resemble the direction of temporarily and spatially local development. 

On the other hand, the PhyCell of \verb|PhyDNet AdvDiff| seems to ignore the direction of $\mathbf u ^{(t)} $ and uses it just to multiply the $\mathbf h_p ^{(t)} $ with the observed precipitation intensities. In this case, $\mathbf u ^{(t)}$ points uniformly South-East, independently of the actual movement directions. Nevertheless, it has to be pointed out that the eastward direction of precipitation movement prevails in the considered Central-European geographical location.

\section{Quadratic Non-linearity}\label{AppQuad}

The first approach, which we call \textit{Quadratic}, is built on a traditional DL paradigm of letting the deep model learn, what is important for loss optimization. $\Phi(\mathbf h_p ^{(t)} )$ from Equation \ref{eq:3_phi} can be described as a first-degree polynomial of spatial partial derivatives. Considering a vector of all $k^2$ partial derivatives $\mathbf{d}(\mathbf h_p ^{(t)})=\left( \mathcal D_{0, 0}(\mathbf h_p ^{(t)}), \dots,\mathcal D_{k-1,k-1}(\mathbf h_p ^{(t)})\right)$ up to some hyperparameter $k$, and vector of learned scalars $\mathbf c=(c_{0,0}, \dots, c_{k-1, k-1})$, the prediction equation may be rewritten as\footnote{The hidden state $\mathbf h_p ^{(t)}$, which is input to the $\Phi$, is omitted here for clarity.}
$	\Phi=\mathbf c \cdot \mathbf{d}.$

We propose to compute all of the possible second-degree terms through matrix multiplication, learn corresponding scalars and add them to this equation. The matrix of second-degree terms $\mathbf d^{(2)}$ is obtained as $
	\mathbf d^{(2)} = UT( \mathbf{d}^\intercal\times\mathbf{d})$, 
where $UT$ is a function selecting only the $\frac{k^2(k^2+1)}{2}$ upper triangular terms
to remove duplicates and flattens them by rows to a row vector. A vector of corresponding scalars $\mathbf{c}^{(2)}$ is learned by $1\times 1$ convolution as in the Section \ref{sec:3_phi} and the prediction equation is extended to
	$\Phi=\mathbf c \cdot \mathbf{d} + \mathbf{c}^{(2)} \cdot \mathbf{d}^{(2)},$
which can be expressed in the expanded form as
\begin{equation}
		\Phi(\mathbf h_p ^{(t)} )
	=\sum_{i,j<k} c_{i,j}\frac{\partial^{i+j}\mathbf h_p}{\partial x^i\partial y^j}(t, \mathbf x) +\sum_{m,n<k;i\leq m;j\leq n} c^{(2)}_{i,j,m,n}\frac{\partial^{i+j}\mathbf h_p}{\partial x^i\partial y^j}(t, \mathbf x)\cdot\frac{\partial^{m+n}\mathbf h_p}{\partial x^m\partial y^n}(t, \mathbf x).
\end{equation}

\end{document}